%% file: LameR-arxiv-v2.tex
\documentclass{article}

\PassOptionsToPackage{numbers, compress}{natbib}

\usepackage[preprint]{neurips_2023}

\usepackage[utf8]{inputenc} 
\usepackage[T1]{fontenc}    
\usepackage{hyperref}       
\usepackage{url}            
\usepackage{booktabs}       
\usepackage{amsfonts}       
\usepackage{nicefrac}       
\usepackage{microtype}      
\usepackage{xcolor}         

\usepackage{wrapfig}
\usepackage{subfigure}
\usepackage{graphicx}

\input{define_operator}
\input{math_commands}

\title{Large Language Models are Strong Zero-Shot Retriever}

\author{%
Tao Shen, Guodong Long, Xiubo Geng, Chongyang Tao, Tianyi Zhou, Daxin Jiang \\
AAII, School of CS, FEIT, UTS,~\texttt{\{tao.shen,guodong.long\}@uts.edu.au} \\
Microsoft Cooperation,~\texttt{\{xigeng,chotao,djiang\}@microsoft.com} \\
University of Maryland,~\texttt{zhou@umiacs.umd.edu}
}

\begin{document}

\maketitle

\begin{abstract}
In this work, we propose a simple method that applies a large language model (LLM) to large-scale retrieval in zero-shot scenarios. Our method, the Language language model as Retriever (LameR), is built upon no other neural models but an LLM, while breaking brute-force combinations of retrievers with LLMs and lifting the performance of zero-shot retrieval to be very competitive on benchmark datasets. Essentially, we propose to augment a query with its potential answers by prompting LLMs with a composition of the query and the query's in-domain candidates. The candidates, regardless of correct or wrong, are obtained by a vanilla retrieval procedure on the target collection. As a part of the prompts, they are likely to help LLM generate more precise answers by pattern imitation or candidate summarization. Even if all the candidates are wrong, the prompts at least make LLM aware of in-collection patterns and genres. Moreover, due to the low performance of a self-supervised retriever, the LLM-based query augmentation becomes less effective as the retriever bottlenecks the whole pipeline. Therefore, we propose to leverage a non-parametric lexicon-based method (e.g., BM25) as the retrieval module to capture query-document overlap in a literal fashion. As such, LameR makes the retrieval procedure transparent to the LLM, thus circumventing the performance bottleneck. 
\end{abstract}


\section{Introduction} \label{sec:intro}

Large-scale retrieval, a.k.a. first-stage retrieval, is to fetch top relevant documents for a given text query from a huge collection with millions to billions of entries. 
It is indispensable in information-seeking tasks, such as open-domain question answering \citep{Chen2017DrQA}, web search \citep{Shen2023LexMAE}, knowledge-grounded dialogue\citep{Zhao2020KGC}, etc. 
Recently, it is also leveraged as a core retrieval-augmenting module to enrich large language models (LLMs) with up-to-date or domain-specific knowledge \citep{Guu2020RetriAugLM,Trivedi2022CotRetri}, which reduces the hallucination problem \citep{Shuster2021RetrievalHalluci} and improves the faithfulness of generated texts \citep{He2023Rethinking}. 
Thereby, large-scale retrieval is a long-term research problem, attracting research efforts from academia and industry.

In the last decade, large-scale retrieval relies heavily on deep representation learning techniques, from bag-of-words (BoW) \citep{Mikolov2013WordEmb} to pre-trained language models (PLMs)  \citep{Devlin2019BERT}. 
Compared to supervised representation learning \citep{Karpukhin2020DPR,Xiong2021ANCE} that requires labor-intensive annotations on query-document pairs, self-supervised (or zero-shot) learning \citep{Lee2019ICT,Ni2021GTR,Izacard2021Contriever,Muennighoff2022SGPT} on in-domain pseudo pairs corpora can be readily generalized to any corpora without human-crafted annotations. Nonetheless, the zero-shot retriever usually results in an inferior retrieval quality \citep{Zhou2022hyperlink}, even worse than a non-parametric term-based BM25 retrieval \citep{Thakur2021BEIR,Zhou2022hyperlink}. 

Fortunately, recent surging LLMs provide a shortcut to reach zero-shot retrieval by augmenting a query with its potential answering elicited from the LLMs \citep{Gao2022hyde}. 
Coupled with a self-supervised retriever, Contriever \citep{Izacard2021Contriever}, it delivers superior retrieval performance, even surpassing a number of supervisedly fine-tuned retrievers. 
But, such a brute-force combination of a self-supervised retriever with a versatile LLM leads to a major problem. The answer elicitation is merely based on prompting LLMs with short, intent-ambiguous, and domain-vague retrieval queries. Due to the ambiguity of user queries and unawareness of in-domain corpora, the LLMs are likely to generate spurious and out-of-domain answers to the queries \citep{Asai2022TaskRetrieval}, making the query augmentation even more toxic.

To circumvent this issue, we propose a brand-new and simple paradigm for large-scale retrieval, called LameR. Essentially, during eliciting LLMs for answers to a query, we inject the query's top answer candidates into the prompt, where the candidates are obtained by applying a vanilla retrieval procedure to the query. As such, the LLMs are prone to distinguish and imitate the candidates \citep{Brown2020GPT3}, while summarizing or/and re-writing new ones with internal knowledge of the LLMs. Moreover, despite correct or wrong candidates, they can at least provide demonstrations about in-domain patterns and knowledge \citep{Min2022ICLRethinking,Xie2022ICLBayesian,Lyu2022ZICL}.

Furthermore, though the LLMs now generate more precise, and reliable query augmentations, the whole pipeline is likely to be bottlenecked by the weak retriever trained on pseudo data in a self-supervised manner. Therefore, we also propose to get rid of any learnable parametric retrievers, while opting for non-parametric term- or lexicon-based retrieval methods (e.g., BM25 in our experiments) in our LameR. In contrast to model-specific compressed and/or latent embeddings from a deep retriever, the lexicon-based retrieval methods capture lexicon overlap between augmented queries and in-collection documents in a literal fashion, thus taking the outputs of LLMs in a transparent mode and bypassing the performance bottleneck problem. 

We evaluate our LameR on several benchmark datasets of large-scale retrieval by following \citet{Gao2022hyde}. Our results show that our proposed method achieves the best retrieval qualities on most datasets compared to other zero-shot competitors. Meantime, it can surpass the LLM-based retriever with in-context labeled demonstrations and outperform the baseline retrievers fine-tuned on full datasets. 

\section{Related Work} \label{sec:relatedwork}

\paragraph{Zero-Shot Large-scale Retrieval. }

In the last years, many research efforts have been dedicated to zero-short retrieval due to its independence of labor-intensive query-document annotations. 
In contrast to zero-shot transfer that supervisedly trains a retriever in one domain and then evaluates it in another domain \citep{Thakur2021BEIR}, we focus on an extremer scenario where no supervised data but the raw target collection is accessible. 
To handle this scenario, previous works construct pseudo query-document pairs from a target retrieval collection, such as  inverse cloze task \citep{Lee2019ICT}, hyperlink prediction \citep{Zhou2022hyperlink}, bottlenecked autoencoder \citep{Shen2023LexMAE}, etc.
Given the mined pseudo pairs, they train a retriever upon pre-trained language models, e.g., BERT and RoBERTa, via contrastive learning with stochastic negatives. 
However, the self-supervised retrievers are only comparable to the lightweight non-parametric term- or lexicon-based retrievers, e.g., BM25 \citep{Robertson2009BM25}.
Even equipped with LLM-based augmentation \citep{Gao2022hyde}, the self-supervised retrievers still lag behind the retrievers fine-tuned on supervised data.
In this work, we discard the inferior self-supervised retrievers but choose the highly generalizable non-parametric retrievers, and propose a brand-new method that integrates LLMs into zero-shot retrieval.

\paragraph{In-context Learning (ICL). }

LLMs can be adapted to new tasks by learning input-label pairs (a.k.a. demonstrations) provided in context, without updates of parameters, which is dubbed in-context learning \citep{Brown2020GPT3}. 
Furthermore, some works seek better in-context demonstrations through retrieval, based on an observation that 
the demonstrations close to the test input help ICL more effectively \citep{Liu2022ICLWhat,Rubin2022ICLL2retrieval}. 
Empirically, ICL, with several demonstrations, remarkably outperforms zero-shot methods across a broad spectrum of tasks, however of a prerequisite for mandatory few-shot examples. 
Fortunately, recent works \citep{Xie2022ICLBayesian,Razeghi2022ICLFreq,Min2022ICLRethinking} suggest ICL demonstrations are mainly used to specify input-label domains and formats of the target task, rather than supervision signal only. 
Sharing a similar inspiration with these works, especially Z-ICL \citep{Lyu2022ZICL}, we leverage a retriever for unsupervised demonstrations from a huge collection to specify the domain, intent, and unit. 
However, we stand with a clean-cut motivation: as we exactly target the retrieval task, the retrieved demonstrations are potential labels (answers), orthogonal to retrieving inputs in previous works \citep{Lyu2022ZICL,Wang2023Query2doc}. 
As such, the demonstrations are likely to help generate correct answers by correction or/and summarization with a boosting inspiration. 

\paragraph{Retrieval \& Rerank Pipeline. }

Our two-stage procedure is similar to the retrieval \& rerank pipeline \citep{Cai2021IRSurvey}. 
The retrieval \& rerank pipeline first employs a high-efficient retriever to fetch top candidates from a collection and then uses a heavy but effective ranker to rerank the candidates for more precise ranking outputs \citep{Gao2022Reranking,Zhou2022r2anker}. 
But, besides requiring supervised data to train both modules, the rerank module is constrained by the upstream retrieval module. In contrast, LameR always lets its retrieval module direct interact with the collection, free of constraint. 




\paragraph{LLM for Information Retrieval. } 
\emph{Although an LLM can directly generate relevant documents and even the final answer for a user query upon its parametric memory, such a generative information-seeking approach is limited} by:
i) out-of-date corpora are learned in the parametric memory, 
ii) unreliable, and hallucinative text is frequently generated, 
and iii) the domain of generated text cannot be specified as demand. 
\emph{In contrast, information retrieval aims to provide in-domain and reliable documents relevant to user queries}, which still dominates people's daily information-seeking methods. Therefore, many research efforts have recently been dedicated to applying large language models (LLMs), such as the GPT series, to information retrieval tasks for superior search performance. The majority of these works are in few-shot or zero-shot scenarios.
\citet{Yu2022GenerateRetrieve}  proposed a generate-then-read pipeline instead of the traditional retrieve-then-read pipeline.
\citet{Dai2022Promptagator} introduced a few-shot dense retrieval approach for different tasks with different retrieval intents. 
\citet{Dua2022AdaptAnnotate} proposed a data augmentation method for domain adaptation for open-domain QA, where a document is passed to LLM for the generation of its possible queries. 
\citet{Gao2022hyde} focused on zero-shot dense retrieval using the Hypothetical Document Embedding (HyDE) method to generate potential answers by LLM as query augmentation. 
\citet{Jeronymo2023InPars-v2} and \citet{Boytsov2023InPars-Light} leveraged a fine-tuned ranker (on MS-MARCO in a supervised manner) to filter LLM-generated data for better query-document quality and thus superior performance. 
\citet{Falcon2023UDAPDR} designed a two-stage LLM pipeline for zero-shot query generation and reranker-distilled retriever. 
\citet{Wang2023Query2doc} utilized a few-shot query-document demonstration to generate documents for a new query as the query's augmentation.

Unlike these works, we focus on the zero-shot retrieval scenario, and neither conduct any in-domain data augmentation for domain-specific retriever training nor introduce any other retrieval or/and intermediate models except for a frozen LLM.

\section{Observations} \label{sec:observe}

In our pilot experiments, we observed the brute-force combination of a versatile LLM with weak retriever leads to certain demerits, which primarily motivates this work. 


\paragraph{Bottleneck by Self-supervised Retriever.}

\begin{wrapfigure}{R}{0.33\textwidth}
\vspace{-22pt}
    \centering
    \includegraphics[width=0.98\linewidth]{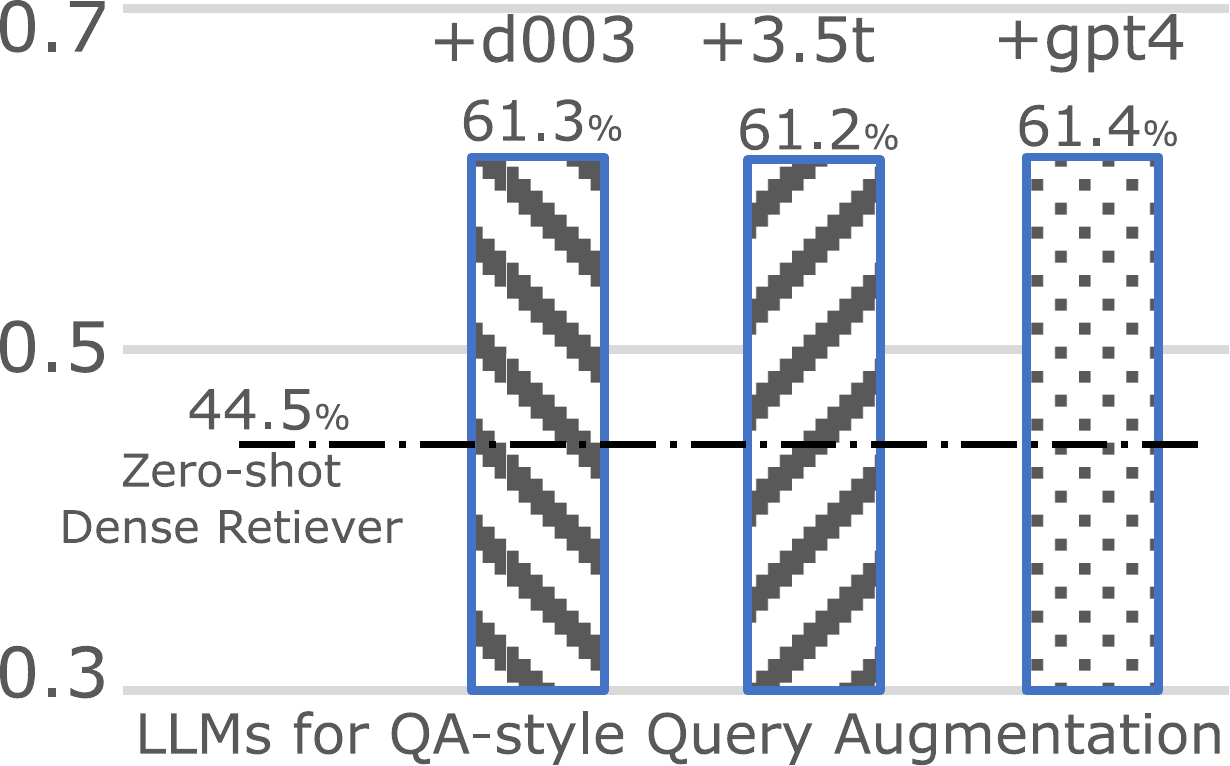}
    \vspace{-4pt}
    \caption{\small nDCG@10 on DL19 for query augmentation w/ LLMs. }
    \label{fig:obs_hyde_llms}
\vspace{-10pt}
\end{wrapfigure}

Due to the weakness of a self-supervised dense retriever in representing capability, the whole pipeline is bottlenecked by the retriever, even though correct answers are likely to be generated by the strong LLM. 
As illustrated in Figure~\ref{fig:obs_hyde_llms}, strengthening LLMs in QA-style query augmentation (i.e., HyDE \citep{Gao2022hyde}, which elicits an LLM to generate answers as query augmentation) hardly improves retrieval performance. Here, `\texttt{d003}' and `\texttt{3.5t}' denote \texttt{text-davinci-003} and \texttt{gpt-3.5-turbo} by OpenAI, respectively. 

    \paragraph{Mismatch w/ Term-based Retriever.}

\begin{wrapfigure}{R}{0.33\textwidth}
\vspace{-14pt}
    \centering
    \includegraphics[width=0.98\linewidth]{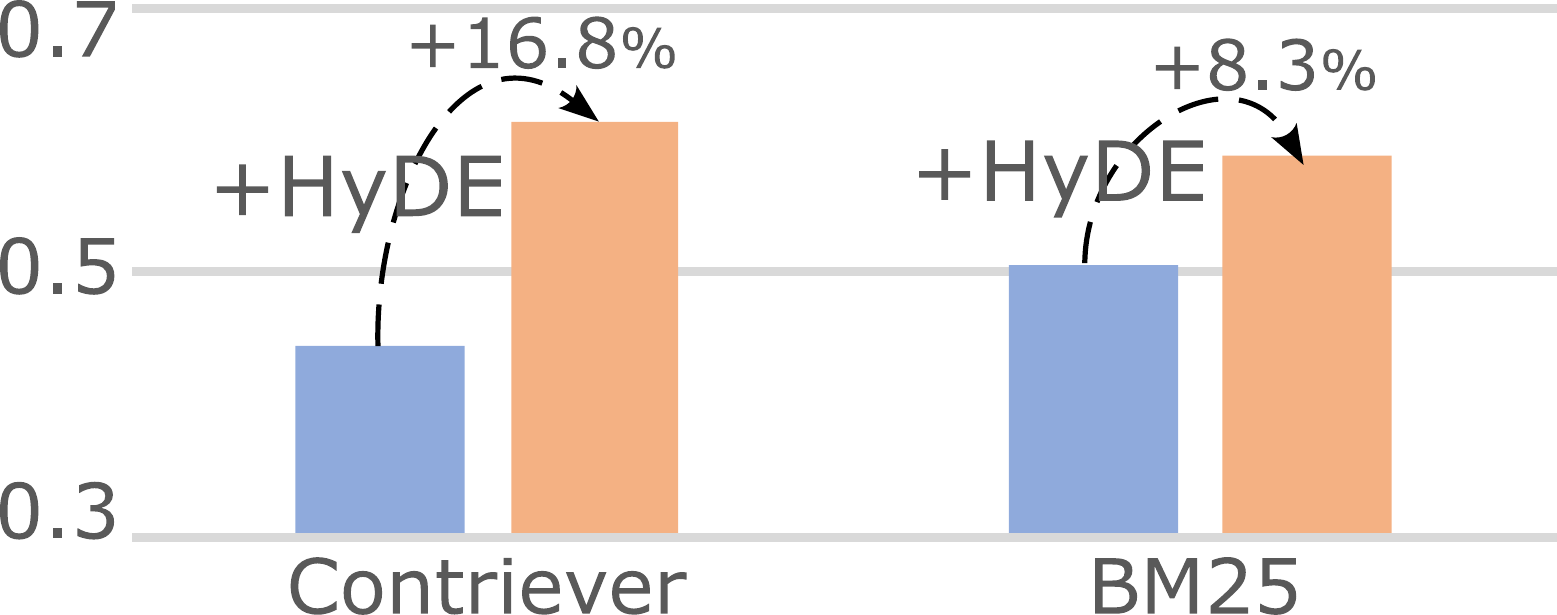}
    \caption{\small HyDE improving Dense and Term-based Retrieval. }
    \label{fig:obs_bm25_mismatch}
\vspace{-10pt}
\end{wrapfigure}

Due to unawareness of in-domain corpora, LLM is likely to generate out-of-domain answers to a given context-short and intent-vague query, making the query augmentation even toxic. 
Thanks to the fuzzy capability of dense retrievers, such query augmentation still bring remarkable improvement in search quality. 
However, when it comes to lexicon-based retrieval (say BM25), the improvement will be reduced due to out-of-domain augmentations.
Quantitatively, as in Figure~\ref{fig:obs_bm25_mismatch},  `Contriever' is a SoTA self-supervised dense retriever while `BM25' is a representative lexicon-based retrieval. It is observed that
although BM25 can beat Contriever in the vanilla setting, HyDE brings twice more improvement to Contriever than BM25, making BM25 less competitive. 

\section{Language Language Model as Retriever (LameR)} \label{sec:methodology}

\begin{figure}
\centering
    \includegraphics[width=0.99\linewidth]{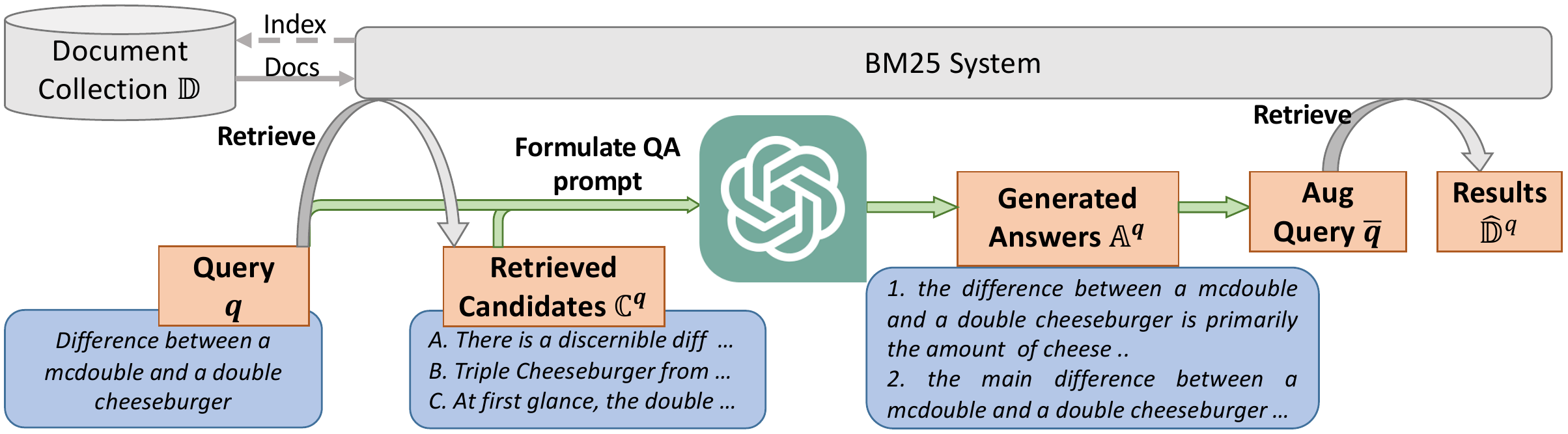}
    \caption{\small \textbf{La}rge language \textbf{m}od\textbf{e}l as \textbf{R}etriever (LameR). Please see Table~\ref{tab:prompt_example} for the prompt formulation. }
    \label{fig:model_illu}
\end{figure}

This section begins with a task definition, followed by elaborations on three components to achieve LameR -- non-parametric lexicon-based retriever (\S\ref{sec:method_retriever}), candidate-prompted answer generation (\S\ref{sec:answer_generation}), and answer-augmented large-scale retriever (\S\ref{sec:ans_aug_retrieval}). LameR's pipeline is illustrated in Figure~\ref{fig:model_illu}.

\paragraph{Task Definition: Zero-Shot Large-Scale Retrieval.}

Providing a huge collection consisting of many documents, $\sD = \{d_i\}_{i=1}^{|\sD|}$, the goal of `large-scale retrieval' is to rank the whole $\sD$ in descending order according to the relevance score between a given text query $q$ and each $d_i$. 
The relevance score is usually derived by a high-efficient retrieval model that operates on a pre-indexed $|\sD|$ and an on-the-fly $q$ to satisfy real-time requirements. 
Meantime, `zero-shot' means that there is no training set with labeled positive query-document pairs for supervised representation learning.

\subsection{Non-parametric Lexicon-based Retriever} \label{sec:method_retriever}

To tackle zero-shot retrieval, a recent trend is to train a deep encoder (e.g., BERT) over pseudo query-document pairs in a self-supervised manner, where the pairs are heuristically mined from the target collection $\sD$. 
Although the self-supervised learning process is required to especially repeat or/and design for every retrieval collection \citep{Lee2019ICT,Zhou2022hyperlink}, the resulting retrieval performance is still not satisfactory in most cases, lagging far behind fully-supervised retrievers. 

In contrast, the non-parametric term- or lexicon-based retrieval methods, e.g., TF-IDF and BM25
\footnote{Although the two hyper-parameters, i.e., $k_1$ and $b$, in BM25 algorithm can be tuned, for example, by grid search, we do not seek to tune them but keep them in defaults, i.e., $k_1 = 0.9$ and $b = 0.4$, in Pyserini \citep{Lin2021pyserini}.}, 
are free of training heavy neural networks, but depend on lexicon overlap with considering term and document frequency of the lexicons. 
Even so, the simple BM25 retrieval method can outperform the self-supervised retriever in many cases in zero-shot retrieval \citep{Zhou2022hyperlink,Thakur2021BEIR}. 

Therefore, in this work we leverage the BM25 method \citep{Robertson2009BM25} to perform large-scale retrieval.
The core idea of BM25 is to rank documents according to their relevance to a given query by incorporating term frequency and inverse document frequency.
In brief, its relevance score between a document $d\in\sD$ and a query $q$ is defined as
\begin{align}
\relevancescore\nolimits^{\text{BM25}}(d, q) \!\!=\!\!\! \sum_{t \in q} \text{IDF}(t) \!\cdot\! \frac{\text{TF}(t, d) \cdot (k_1 + 1)}{\text{TF}(t, d) \!+\! k_1 \!\cdot\! (1 \!-\! b \!+\! b \!\cdot\! \frac{\text{len}(d)}{\text{avgdl}})},~\text{where}~\text{IDF}(t) \!\!=\!\! \log \frac{N \!-\! n(t) \!+\! 0.5}{n(t) + 0.5}.
\end{align}
Here, $t$ denotes a lexicon term in $q$, $\text{TF}(t, d)$ is the term frequency of $t$ in document $d$, and $\text{IDF}(t)$ is the inverse document frequency of term $t$, $N=|\sD|$ is the total number of documents in the collection, $n(t)$ is the number of documents containing term $t$, $\text{len}(d)$ is the length of $d$, and $\text{avgdl}$ is the average document length across the collection. 
In the remainder, we define a retrieval procedure as
\begin{align}
    \hat\sD^q = \retriever(q, \sD, K).
\end{align}
$\hat\sD^q$ is a list of top-K retrieval candidates of $q$
in descending order w.r.t relevance scores, so $|\hat\sD^q|=K$.

\begin{table*}[t]
\centering
\small
\caption{\small Our simple QA prompt to elicit knowledge from LLM for information retrieval in our LameR. Here, the entry with `\{$\cdot$\}' represents a placeholder for the corresponding text. $c^q_l\in\sC^q$ denotes a retrieved candidate. Please see Appendix~\ref{sec:prompts} for the prompts for all datasets. }
\label{tab:prompt_example}
\begin{tabular}{p{12cm}}
\toprule
\textbf{Candidate-prompted Instruction}. \\ \midrule
\texttt{Give a question ``\{$q$\}'' and its possible answering passages (most of these passages are wrong) enumerated as: \textbackslash{}n~1.\{$c^q_1$\} \textbackslash{}n 2.\{$c^q_2$\} \textbackslash{}n 3.\{$c^q_3$\} $\dots$} \\ \texttt{please write a correct answering passage.} \\
\bottomrule
\end{tabular}
\end{table*}

\paragraph{Remark.} 
When employing a strong, non-tunable, generative model, e.g., LLM, for explicit text augmentations of a query, 
a lexicon-based retrieval method has its own merit in not only high efficiency, but taking the exact augmentations for retrieval without compressed embedding. 
Therefore, using the lexicon-based method exposes LLMs' outputs to the retrieval collection literally, making the retrieval module transparent to LLMs.
By comparison, the neural encoder, trained on heuristically mined pseudo data in self-supervised, is too weak to model the LLM-augmented queries, leaving a performance bottleneck here (see \S\ref{sec:observe}). 

\subsection{Candidate-Prompted Answer Generation} \label{sec:answer_generation}

Given a query $q$, we augment it with its answer(s) $a$ elicited from an LLM, which has been proven effective in improving zero-shot retrieval quality \citep{Gao2022hyde,Wang2023Query2doc}. How to conduct the elicitation remains an open question. 
For example, in a straightforward way, \citep{Gao2022hyde} propose to prompt an LLM with a composition of a QA instruction and the query. 
However, as the LLM can only receive a short, intent-ambiguous query, joined with a broad and general QA instruction, it is not well instructed by the prompt with both the {intent} and {domain} of a query, leading to less precise answers. 
\citep{Wang2023Query2doc} add few-shot query-document examples as in-context demonstrations to the prompt for more reasonable answers, which, however, is unavailable in zero-shot settings. 

Instead, we propose a new prompt schema, called candidate-prompted answer generation, for query augmentation in large-scale retrieval. 
As shown in Table~\ref{tab:prompt_example}, besides a task instruction and a retrieval query, a list of top answering candidates is also included in the prompt for elicitation of an LLM. 
Here, the top candidates are obtained by directly applying a vanilla retrieval process to the query via the retriever (\S\ref{sec:method_retriever}).
Formally, we first retrieve top-$M$ candidates for $q$ from the whole $\sD$ by
\begin{align}
    \sC^q = \retriever(q, \sD, M), \label{equ:candi_gen}
\end{align}
where $M$ is usually very small (e.g., $<10$) to reduce computation overhead for downstream modules. 
Then, to elicit knowledge from an LLM, we construct a prompt with $\sC^q$ and then invoke the LLM for answer generation, i.e., 
\begin{align}
    \sA^q = \{a^q_1 \dots a^q_N| a^q\sim \llm\left(p\left(t, q, \sC^q\right)\right) \}  \label{equ:ans_gen}
\end{align}
where $p(\cdot)$ composes the prompt using task an instruction $t$, the query $q$, and the retrieved candidates $\sC^q$ (see Table~\ref{tab:prompt_example} for an example and Appendix~\ref{sec:prompts} for prompts of all tasks).
It is noteworthy that we generate multiple (i.e., $N$) answers by sampling outputs of the LLM, because we'd like to provide as many potential answers as we can to prevent the `vocabulary mismatch' problem.

As such, $\llm(\cdot)$ utilizes the answering candidates $\sC^q$ in two aspects: 
i) If one or many gold documents of $q$ existing in $\sC^q$, $\llm(\cdot)$ serves like a re-ranker and generates the answers $\sA^q$ by both summarizing the correct documents from $\sC^q$ and eliciting internal parameterized knowledge. 
ii) Regardless of the correctness of $\sC^q$, $\llm(\cdot)$ also receives in-collection answering information about intents, domains, and units, which are prone to help the LLM generate more precise answers $\sA^q$.

\subsection{Answer-Augmented Large-Scale Retrieval} \label{sec:ans_aug_retrieval}

Given the generated answers $\sA^q$ of $q$, we use them to augment $q$ and produce a new query $\bar q$.
Attributed to the non-parametric lexicon-based retriever, we can perform the query augmentation in a very straightforward way, which operates on plain text rather than latent embeddings. 
That is, we can easily concatenate every $a^q\in\sA^q$ with the original $q$, i.e.,
\begin{align}
    \bar q = \concat(q, a^q_1, q, a^q_2, \dots, q, a^q_N), \label{eq:query_aug}
\end{align}
where $\concat$ denotes a concatenation operation in text. 
Lastly, we simply use the augmented query, $\bar q$, to conduct a large-scale retrieval, 
\begin{align}
    \hat\sD^{\bar q} = \retriever(\bar q, \sD, K), \label{eq:query_aug_retrieval}
\end{align}
where $\hat\sD^{\bar q}$ is a list of final retrieved documents for query $q$ and $K=1000$ for metric calculation. 
Thanks to the high efficiency of the lexicon-based retriever with an inverted index, the augmentation would not cause catastrophic overhead increases, which is still faster than a dense retriever.

\section{Experiment}

In this section, we will conduct extensive experimental evaluations of the proposed retrieval method and compare it with strong competitors. 

\paragraph{Datasets and Metrics.} 
Following the datasets used by \citet{Gao2022hyde}, we first employ the widely-used passage retrieval datasets, MS-MARCO~\citep{Nguyen2016MSMARCO} and report performance on TREC Deep Learning 2019 \citep{Craswell2020TREC19} and TREC Deep Learning 2020\citep{Craswell2021TREC20} test sets (DL19 and DL20 for short, respectively). 
Meantime, we also evaluate our method on BEIR benchmark~\citep{Thakur2021BEIR}. 
Here, we follow \citet{Gao2022hyde} to consider low-resource datasets from the BEIR dataset, so we employ six datasets, consisting of one fact-checking task (Scifact), one question-answering task (FiQA), one bio-medical IR task (TREC-COVID), one news retrieval task (TREC-NEWS), one argument retrieval task (ArguAna), and one entity retrieval task (DBPedia).
Note that, as a zero-shot retrieval setting, we do not use any training query-document pairs but directly evaluate our method in the test sets. 
Following previous works, we report MAP, nDCG@10 and Recall@1000 (R@1k) for both TREC Deep Learning 2019 and TREC Deep Learning 2020. And nDCG@10 is reported for all the datasets in the BEIR benchmark.

\paragraph{Experimental Setup.}
As for the large language model, we use \texttt{gpt-3.5-turbo} as the LLM to perform answer generation by default. Meantime, we also involve \texttt{gpt-4} to investigate whether stronger LLM will bring more improvement. 
And, the number of candidates, $M$ in Eq.(\ref{equ:candi_gen}), is set to 10 in our main results, and the number of generated answers, $N$ in Eq.(\ref{equ:ans_gen}) is set to 5. 
To ensure efficiency, we truncate each of the queries and passages/documents to 128 tokens.

\paragraph{Baselines and Competitors. }
As we focus on the zero-shot retrieval setting, our main baselines fall into the retrieval methods without dependency on annotated query-document pairs (i.e., \textit{w/o} relevance judgment). 
In particular, we use BM25 \citep{Robertson2009BM25} and Contriever \citep{Izacard2021Contriever} as strong baselines for zero-shot lexicon and dense retrieval, respectively. And, we also include HyDE \citep{Gao2022hyde} as the state-of-the-art competitor for LLM-based retrieval. 
Furthermore, we also employ some baselines not in zero-shot settings to verify the effectiveness of our method. 
On the one hand, we leverage Q2D+BM25 \citep{Wang2023Query2doc} as a few-shot baseline (i.e., \textit{w/ few-shot relevance judgment}), where in-context gold query-document pairs are provided to help LLM generate answers for a query. 
On the other hand, we consider some popular fully-supervised retrieval models (i.e., \textit{w/ relevance judgment}), including DPR \citep{Karpukhin2020DPR}, ANCE \citep{Xiong2021ANCE}, fine-tuned Contriever \citep{Izacard2021Contriever}, etc.

\subsection{Main Evaluation}

\begin{table*}[t]
\centering
\small
\caption{\small Results for web search on DL19/20. 
Best performing w/o relevance judgment is marked \textbf{bold}. 
DPR, ANCE and Contriever$^\text{FT}$ are in-domain \emph{supervised} models that are finetuned on MS-MARCO training data.}
\label{tab:main_dl}
\begin{tabular}{l|ccc|ccc}
\toprule
 & \multicolumn{3}{c|}{\textbf{TREC Deep Leaning 2019}}          & \multicolumn{3}{c}{\textbf{TREC Deep Leaning 2020}} \\
 & MAP  & nDCG@10 & R@1k & MAP  & nDCG@10 & R@1k \\
\midrule
\multicolumn{7}{l}{\textit{w/o relevance judgment (zero-shot retrieval)}} \\
BM25 & 30.1 & 50.6 & 75.0 & 28.6 & 48.0 & 78.6 \\
Contriever & 24.0 & 44.5 & 74.6 & 24.0 & 42.1 & 75.4 \\
HyDE & {41.8} & {61.3} & {88.0}	& {38.2}	& {57.9} & {84.4} \\
\textbf{LameR (ours)} &\textbf{47.2} & \textbf{69.1}& \textbf{89.9}& \textbf{45.6}& \textbf{64.8}& \textbf{88.7}\\
\midrule
\multicolumn{7}{l}{\textit{w/ few-shot relevance judgment  (few-shot ICL for answer generation)}} \\
Q2D$_\text{BM25}$ & - & 66.2 & - & - & 62.9 &  - \\
\midrule
\multicolumn{7}{l}{\textit{w/ relevance judgment (fully-supervised fine-tuning)}} \\
DPR  & 36.5 & 62.2 & 76.9 & 41.8 & {65.3} & 81.4 \\
ANCE & 37.1 & {64.5} & 75.5 & 40.8 & 64.6 & 77.6 \\
Contriever$^\text{FT}$ & 41.7 & 62.1 & 83.6 & {43.6} & 63.2 & {85.8} \\
\bottomrule
\end{tabular}
\end{table*}

\begin{table}[t]
\centering
\small
\caption{\small Low resource tasks from BEIR. 
Best performing w/o relevance judgment are marked \textbf{bold}.
}\label{tab:main_beir}
\begin{tabular}{l|cccccc}
\toprule
nDCG@10           & Scifact & Arguana & Trec-COVID & FiQA    & DBPedia & TREC-NEWS \\
\midrule
\multicolumn{7}{l}{\textit{w/o relevance judgment}} \\
BM25       & 67.9    & 39.7   & {59.5}     &  23.6 &  31.8   & 39.5  \\
Contriever & 64.9    & 37.9   & 27.3     &  24.5 &  29.2  &  34.8  \\
HyDE      & {69.1}    & \textbf{46.6}   & 59.3     &  \textbf{27.3} &  {36.8}  &  {44.0}  \\
\textbf{LameR (ours)} & \textbf{73.5} & 40.2 & \textbf{75.8} & 25.8 & \textbf{39.0} & \textbf{50.3} \\
\midrule
\multicolumn{7}{l}{\textit{w/ few-shot relevance judgment}} \\
Q2D$_\text{BM25}$        & 68.6    & -    & 72.2      & -   & 37.0    & - \\
\midrule
\multicolumn{7}{l}{\textit{w/ relevance judgment}} \\
DPR        & 31.8    & 17.5    & 33.2      & 29.5    & 26.3    & 16.1 \\
ANCE       & 50.7    & 41.5    & {65.4}      & 30.0    & 28.1    & 38.2 \\
Contriever$^\text{FT}$ & 67.7 & 44.6 & 59.6 & {32.9} & {41.3} & 42.8 \\
\bottomrule
\end{tabular}
\end{table}

\paragraph{DL19 and DL20 Test Sets.}
As shown in Table~\ref{tab:main_dl}, we compare our LameR with its baselines and competitors in both TREC Deep Learning 2019 and 2020 test sets. 
It is observed that our method achieves the best performance in the zero-shot setting, significantly outperforming its strong competitor, HyDE\footnote{Although HyDE uses \texttt{text-davinci-003} as its LLM, we found updating it with \texttt{gpt-3.5-turbo} leads to similar retrieval performance. See Figure~\ref{fig:obs_hyde_llms} for details.}. 
This clearly verifies the effectiveness of our candidate-prompted answer generation. It is also noteworthy that our LameR is based on a much faster BM25 retriever, in contrast to the heavy dense retriever, Contriever, in HyDE. 
Meantime, compared to the method (Q2D$_\text{BM25}$) with few-shot relevance judgment and the methods (DPR, etc.) with full relevance judgment, our proposed LameR achieves the best on most retrieval evaluation metrics.

\paragraph{BEIR Benchmark. } 
Furthermore, we compare our retrieval method with the others on six low-resource tasks from the BEIR dataset. As shown in Table~\ref{tab:main_beir}, our proposed method performs best on four out of six datasets. 
It should be highlighted that our LameR achieves superior performance on two TREC retrieval datasets, i.e., TREC-COVID and TREC-NEWS, which verify our proposed method in web information-seeking tasks. 
Meantime, We found our LameR delivers poor results on `Argunan', a dataset designed to retrieve counter-argument passages from a collection. Since the queries and documents in the dataset are usually over-long ($>$ 256), this is possibly caused by applying aggressive truncation (cap at 128) to the long queries and passages in the dataset. 
Besides, we also noticed that the performance of FiQA in zero-shot settings is far from that in the few-shot or fully-supervised settings. This may be caused by the lack of financial knowledge in general LLM. 

\subsection{Ablation Study and Further Analysis}

\begin{figure}[t]
\subfigure[]{
\includegraphics[width=0.27\linewidth]{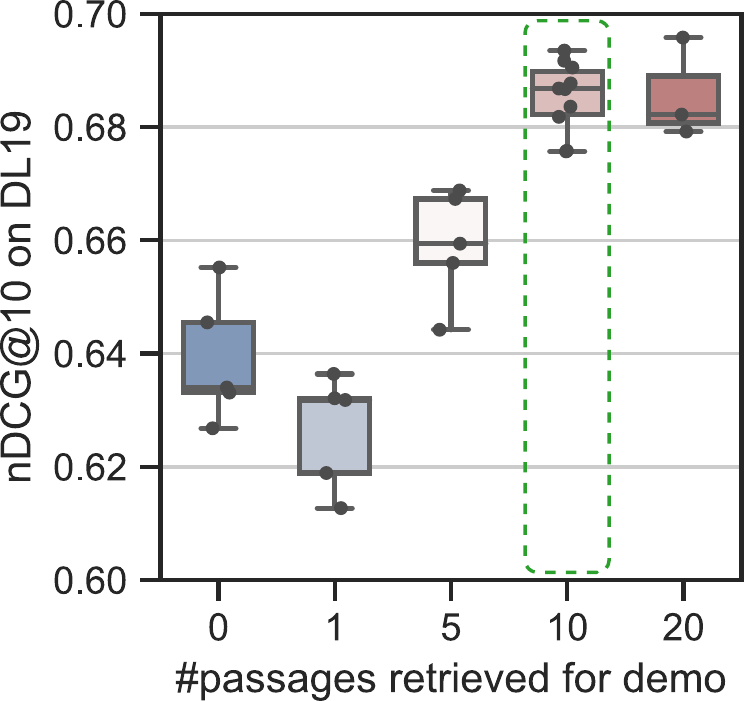}
\label{fig:boxplot_retri_num}}
\subfigure[]{
\includegraphics[width=0.22\linewidth]{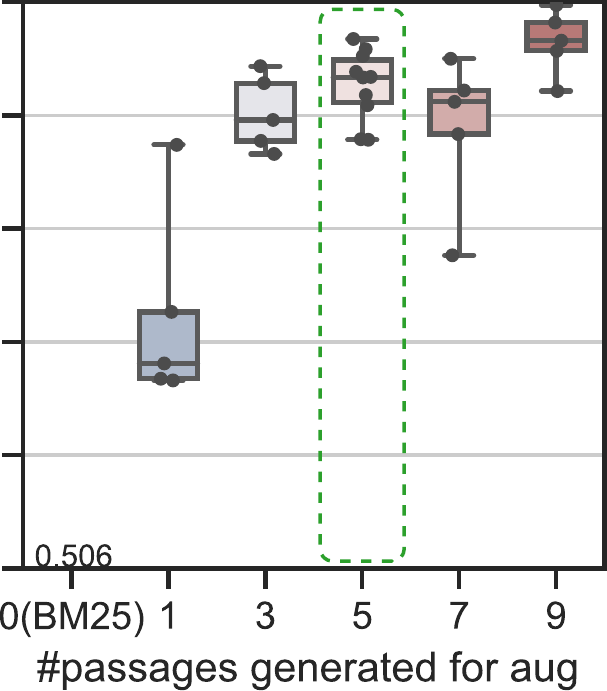}
\label{fig:boxplot_gen_num}}
\subfigure[]{
\includegraphics[width=0.22\linewidth]{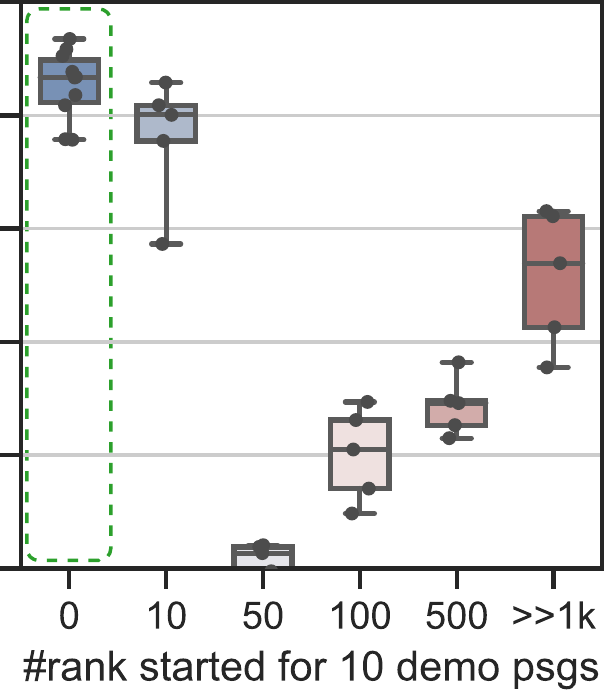}
\label{fig:boxplot_start_point}}
\subfigure[]{
\includegraphics[width=0.224\linewidth]{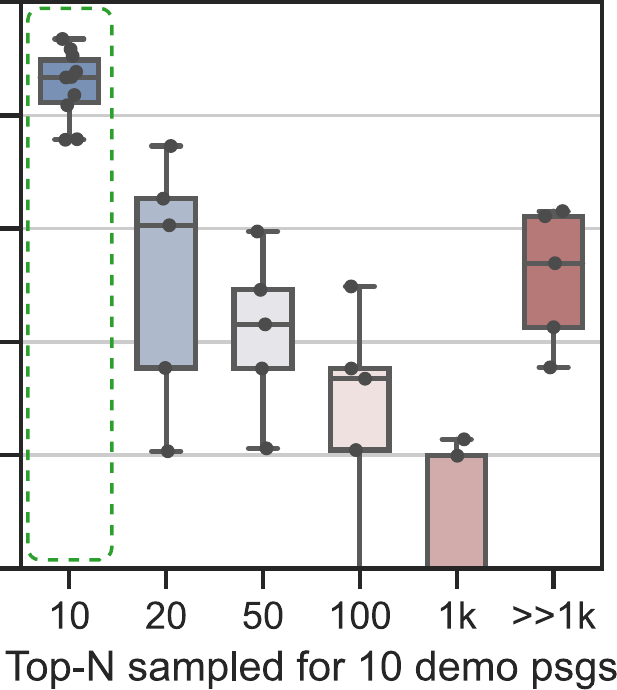}
\label{fig:boxplot_top_sample}}
\vspace{-12pt}
\caption{\small
Hyperparameter explorations and ablation studies, where the data points in dashed rectangles denote our default choices. 
\textbf{(a)} The number of retrieved passages as in-context demonstration for answer generation, i.e., $M$ in Eq.(\ref{equ:candi_gen}). 
\textbf{(b)} The number of generated answers as query augmentations for large-scale retrieval, i.e., $N$ in Eq.(\ref{equ:ans_gen}). 
\textbf{(c)} and \textbf{(d)} depict the schemes to obtain the 10 demo-passages, where the first is to fetch 10 consecutive passages from a \textit{start index} of the BM25-retrieved passages and the second is to randomly sample 10 passages from \textit{top-N} passages. Note that `$\gg$1k' denotes randomly sampling 10 passages from the whole collection. 
}\label{fig:boxplot}
\end{figure}

\paragraph{Number of Retrieved Demos.}
First, we investigate whether the number of retrieved passages (as in-context demonstration) affects query augmentation and thus retrieval quality. 
As shown in Figure~\ref{fig:boxplot_retri_num}, increasing $M>0$ consistently brings improvement in answer-augmented large-scale retrieval, and the improvement becomes marginal when the number exceeds 10. 
Considering that increasing $M$ inevitably causes more computation overheads, we use $M=10$ for a better trade-off between performance and efficiency.
Besides, an interesting point is that LameR with $M=0$ is surprisingly better than both
i) HyDE, which verifies the effectiveness of our query augmentation coupled with BM25 retrieval, i.e., Eq.(\ref{eq:query_aug}-\ref{eq:query_aug_retrieval}),
and ii) LameR with $M=1$, which is likely caused by low recall performance in top-1 and more severe interference of error candidates.

\paragraph{Number of Answer Generations.}
We also investigate whether the number of answers generated by LLM will affect the performance of our LameR. 
As shown in Figure~\ref{fig:boxplot_gen_num}, the performance of retrieval grows along with the number of generated answers, but becomes fluctuating and saturated when $N>5$. 
Therefore, we use $N=5$ as the default in our experiments.

\paragraph{Schemes to Obtain Demo-passages.}

It is \textit{de facto} to leverage top-10 retrieved passages as demonstrations as they are likely to provide pivot query-related knowledge in a limited context window of LLMs. 
To empirically check this intuition, we propose three schemes for demo-passages: 
i) As shown in Figure~\ref{fig:boxplot_top_sample}, the performance consistently drops when we increase the sample range because the related knowledge and correct demonstrations are weakened gradually.
ii) As shown in Figure~\ref{fig:boxplot_start_point}, we fetch 10 consecutive passages from different start indices in BM25 results. Surprisingly, there is a U-shaped curve, which can be explained by `hard negatives' widely presenting in IR: 
Basically, hard negatives in top candidates challenge LLMs' distinguishing capability between positives and hard negatives. What's worse, with increasing start indices, the correct passages scarcely appear in the 10 consecutive passages, making the LLMs lose contrastive samples and get fooled by the negatives. 
iii) More interestingly, as the `$\gg$1k' in both Figure~\ref{fig:boxplot_start_point} \& \ref{fig:boxplot_top_sample}, randomly sampling 10 entries from the whole collection as demo-passages results in surprisingly high results. 
This is because they are focused on providing useful information about the knowledge domain (e.g., web, news, Wikipedia, scientific, arguments), task intent (e.g., dialogue, question answering), answering format (e.g., unit, length, pattern), etc., while free from hard negatives or spurious answers.


\paragraph{Exploring Extremes of LameR.}

\begin{wraptable}{r}{0.42\textwidth} 
\vspace{-15pt}
\small 
\centering
\caption{\small Exploring extremes of LameR.}\label{tab:extremes}
\setlength{\tabcolsep}{3pt}
\begin{tabular}{l|ccc}
\toprule
\textbf{DL19} & MAP  & nDCG@10 & R@1k \\
\midrule
BM25 & 30.1 & 50.6 & 75.0 \\
\textbf{LameR} (dflt) & 47.2 & 69.1 & 89.9 \\
LameR-oracle & 60.7 & 84.0 & 93.8 \\
\midrule
$\Diamond$ 2nd Round & 46.7 & 68.1 & 87.5 \\
\bottomrule
\end{tabular}
\vspace{-13pt}
\end{wraptable}

As LameR is built upon BM25 retrieval system, the lower bound of LameR would be BM25. 
Go beyond, it is interesting to find out the upper bound of LameR, which can demonstrate the extreme performance that LameR may deliver. 
As shown in Table~\ref{tab:extremes}, we conduct an experiment called `LameR-oracle', where 10 demo-passages are instead obtained by gold query-document pairs in the labeled test set. 
It's seen that compared to our LameR w/ default settings (i.e., dflt), LameR-oracle performs much higher, verifying i) the importance of the correctness of demonstrated passages and ii) a great improvement room left for further research. 
As an initial exploration, we propose a brute-force attempt that a 2nd-round LameR is applied to the retrieval results by default LameR, but to our surprise, the performance even drops by absolute 1.0\% nDCG@10 (see the last row of Table~\ref{tab:extremes}). 
Sharing inspirations with error reinforcement, 
the query augmented by an LLM (in the 1st round) is prone to return spurious passages that especially confuse the LLM (i.e., hardly distinguished), resulting in wrong answers to poison BM25. This suggests that in the future, we should focus more on introducing multiple retrieval methods to achieve diversity.

\paragraph{Power of Stronger LLM.}


\begin{wraptable}{r}{0.3\textwidth} 
\vspace{-23pt}
\small 
\centering
\caption{\small LameR with GPT4. }\label{tab:strong_llm}
\begin{tabular}{l|c}
\toprule
\textbf{DL20}  & nDCG@10 \\
\hline
BM25  & 48.0 \\
HyDE	& {57.9} \\
DPR (supv.) & 65.3 \\
\hline
\textbf{LameR$_{\text{GPT-3.5}}$} & {64.8} \\
\textbf{LameR$_{\text{GPT-4}}$} & \textbf{65.9} \\
\bottomrule
\end{tabular}
\vspace{-18pt}
\end{wraptable}

To further verify if our LameR will benefit from stronger LLM, we involve the bleeding-edge LLM, GPT-4, in our LameR framework and apply it to DL20 dataset as its results in the main evaluation with GPT-3.5 is not superior enough. 
As shown in Table~\ref{tab:strong_llm}, after applying GPT-4, our retrieval method achieves significantly high performance and beats all the competitors even with full relevance judgment.

\subsection{Efficiency Analysis} \label{sec:efficiency_analysis}

\begin{wrapfigure}{R}{0.5\textwidth}
\vspace{-42pt}
    \centering
    \includegraphics[width=0.98\linewidth]{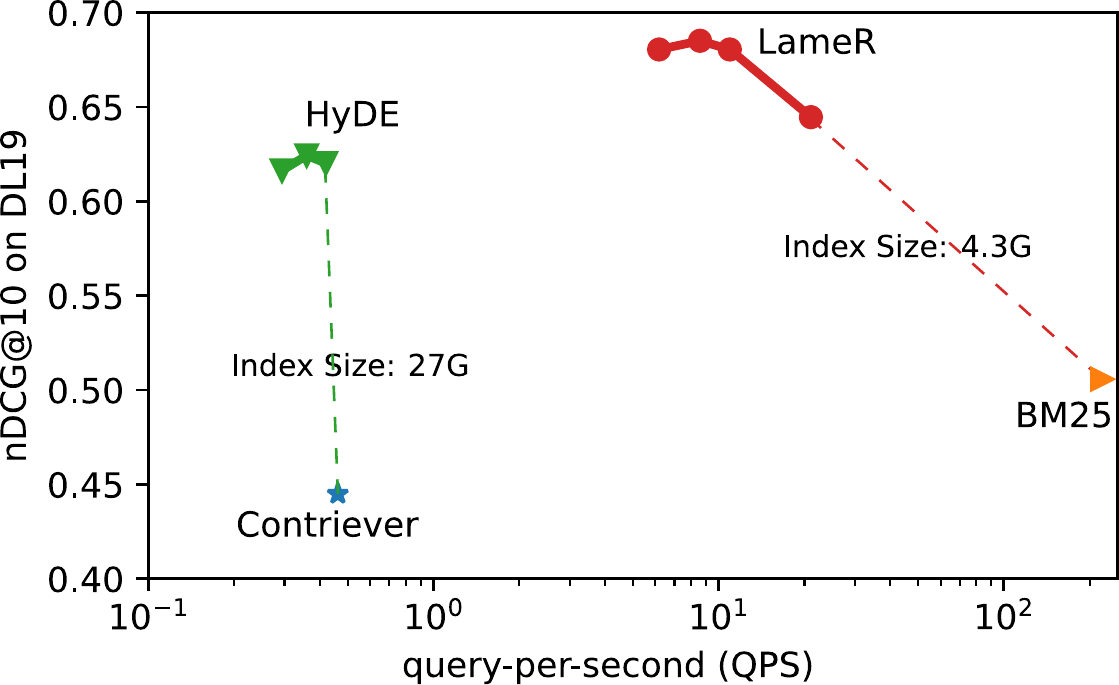}
    \vspace{-4pt}
    \caption{\small Efficiency of LameR with HyDE in retrieval latency (QPS) and index size (GB). Numbers for LameR sum overheads in two stages, and the variants for each system are achieved by changing generation number. }
    \label{fig:efficiency}
\vspace{-10pt}
\end{wrapfigure}

\paragraph{Overheads w/ LLMs.}
Similar to HyDE~\citep{Gao2022hyde} and Q2D~\citep{Wang2023Query2doc}, using LLMs to generate query augmentations inevitably leads to high computation overheads. 
Optimistically speaking, such inference-only overheads do not increase with the scale of retrieval collection, and a recent trend is to make smaller LLMs competitive \citep{Touvron2023llama,Taori2023alpaca}, which would benefit these methods. In the future, we will explore specializing in a smaller LLM to generate query augmentations. 
Besides, in HyDE and our LameR, introducing LLMs makes the whole retrieval system free from heavy query-document annotations and outperforms fully-supervised baselines. 
Specifically, as few-shot Q2D and our zero-shot LameR use extra passages in contrast to zero-shot HyDE, they outperform HyDE significantly. 
Comparing zero-shot LameR with few-shot Q2D, with similar LLM's overheads (i.e., reducing our retrieved candidates), the LameR achieves 66.7\% nDCG@10 on DL19, still surpassing Q2D.

\paragraph{Overheads in Retrieval.}
Moving to overheads in retrieval, we compare BM25-based zero-shot LameR with its counterpart, HyDE, equipped with zero-shot dense retriever. 
As in Figure~\ref{fig:efficiency}, benefiting from highly-efficient BM25, LameR, with much higher zero-shot retrieval performance, wins in both retrieval latency and index size.

\subsection{LameR meets Dense Retriever}

Given promising results w/ a simple BM25, we explore replacing the 2nd-stage BM25 w/ an encoder for dense retrieval. Compared to Eq.(\ref{eq:query_aug}), the dense embedding of an augmented query is derived by $\bar\vq = 1/N \cdot \sum\nolimits_{l\in[1,N]} (\enc(q;\theta^\text{(den)}) + \enc(a^q_l);\theta^\text{(den)})/2$, where $\theta^\text{(den)}$ parameterizes $\enc(\cdot)$. 

\begin{table*}[t]
\centering
\small
\caption{\small Results on DL19/20. 
$\dag$Equipping with our implemented SimLM \citep{Wang2022simlm}.
We mark the `absolute improvement over base retriever' in superscript for key methods.  
Ref: DPR~\citep{Karpukhin2020DPR}, SimLM~\citep{Wang2022simlm}, and E5~\citep{Wang2022e5}.
}
\label{tab:lamer_dense}
\begin{tabular}{l|ccc|ccc}
\toprule
 & \multicolumn{3}{c|}{\textbf{TREC Deep Leaning Track 2019}}          & \multicolumn{3}{c}{\textbf{TREC Deep Leaning Track 2020}} \\
 & MAP  & nDCG@10 & R@1k & MAP  & nDCG@10 & R@1k \\
\midrule
\multicolumn{7}{l}{\textit{Zero-shot Retriever}} \\
BM25 & 30.1 & 50.6 & 75.0 & 28.6 & 48.0 & 78.6 \\
\textbf{LameR}$_{\text{bm25}}$ (zero-shot) &{47.2} & ~~~~~~{69.1}$^\text{+18.5}$& {89.9}& {45.6}& ~~~~~~{64.8}$^\text{+16.8}$& {88.7}\\
\hline
Contriever & 24.0 & 44.5 & 74.6 & 24.0 & 42.1 & 75.4 \\
\textbf{LameR}$_{\text{Contriever}}$ (zero-shot) &{41.1} & ~~~~~~{64.3}$^\text{+19.8}$ & {87.3}& 38.3 & ~~~~~~58.2$^\text{+16.1}$ & 85.5\\
\midrule
\hline
\multicolumn{7}{l}{\textit{Fully-supervised Retriever}} \\
Contriever$^\text{FT}$ & 41.7 & 62.1 & 83.6 & {43.6} & 63.2 & {85.8} \\
DPR  & 36.5 & 62.2 & 76.9 & 41.8 & {65.3} & 81.4 \\
SimLM & - & 71.4 & - & - & 69.7 & - \\
E5$_{\text{base}}$ & - & 74.3 & - & - & 70.7 & - \\
\midrule
\multicolumn{7}{l}{\textit{LLM-augmented Fully-supervised Retriever}} \\
HyDE$_{\text{Contriever}^\text{FT}}$ & - & 67.4 & - & - & 63.5 & - \\
Q2D$_\text{DPR}$ & - & 68.7 & - & - & 67.1 & - \\
Q2D$_\text{SimLM}$ & - & ~~~~~{72.9}$^\text{+1.5}$ & - & - & ~~~~~{71.6}$^\text{+1.9}$ & - \\
Q2D$_{\text{E5}_\text{base}}$ & - & ~~~~~{74.9}$^\text{+0.6}$ & - & - & ~~~~~{72.5}$^\text{+1.8}$ & - \\
\textbf{LameR}$_{\text{SimLM}}$$\dag$ & \textbf{54.9} & ~~~~~\textbf{76.5}$^\text{+5.1}$ & \textbf{91.1} & \textbf{55.7} & ~~~~~\textbf{75.8}$^\text{+6.1}$ & \textbf{89.5} \\
\bottomrule
\end{tabular}
\vspace{-16pt}
\end{table*}

\paragraph{Consistency across Paradigms.}
Recall the results in \S\ref{sec:observe}: Applying HyDE leads to inconsistent improvement on zero-shot dense retrieval (i.e., Contriever) and term-based retriever (i.e., BM25). So, we'd like to check if LameR can overcome this issue by considering in-domain demonstrations. 
As listed in Table~\ref{tab:lamer_dense}(top), applying LameR to Contriever and BM25 results in similar improvement, verifying its effectiveness in query augmentation by demonstrating in-domain knowledge.

\paragraph{LameR w/ SoTA Retriever.}
To exploit the performance extreme of LameR, we incorporate a SoTA dense retriever, SimLM \citep{Wang2022simlm}. 
As shown in Table~\ref{tab:lamer_dense}(bottom), {LameR}$_{\text{SimLM}}$ significantly improves the SoTA performance on DL19 and DL20 and achieves the best effectiveness. 
Meantime, compared to {Q2D}$_{\text{SimLM}}$, our LameR brings significantly higher improvement to SimLM than Q2D (by 3.6\% and 4.2\% on DL19 and DL20, respectively), not to mention Q2D relying on few-shot demonstration.

\section{Limitations.}
i) \textit{Instruction sensitivity}: Identical to other prompt-based LLM applications, this work would also be sensitive to the instructions with different LLMs, which may consume a lot of human effort on prompt writing. 
ii) \textit{Computation Overheads}: As stated in \ref{sec:efficiency_analysis}, although the 2-stage retrieval procedure in LameR is very fast by inheriting BM25, LameR is constrained by calling the LLM for answer generation in terms of computation overheads. 
To overcome these limitations, in the future we will explore specializing in a relatively smaller LLM for query-augmentation purposes.

\section{Conclusion}
In this work, we propose a retrieval method based merely on a large language model (LLM) and a simple BM25 algorithm, without any dependence on learnable retrieval models. 
As such, all the operations are performed in the consistent interface of natural language (i.e., language-based query augmentation and lexicon-overlap retrieval relevance), without the performance bottleneck of a fragile self-supervised model-based retriever.
The extensive experimental evaluations verify the effectiveness of the proposed LameR, supporting that the large language model can solely serve as a strong retriever without any in-domain annotated query-document pairs.


\bibliography{ref,ref_urls}
\bibliographystyle{plainnat}
\appendix

\section{All Prompts} \label{sec:prompts}
We did not carefully craft the prompts in this work but directly adapted the prompts in \citep{Gao2022hyde}. We write our prompts of LameR for all the datasets in Table~\ref{tab:prompt_all}.

\begin{table*}[t]
\centering
\small
\begin{tabular}{p{12cm}}
\toprule
Prompt for \textbf{DL19} and \textbf{DL20}. \\ \midrule
\texttt{Give a question ``\{$q$\}'' and its possible answering passages (most of these passages are wrong) enumerated as: \textbackslash{}n~1.\{$c^q_1$\} \textbackslash{}n 2.\{$c^q_2$\} \textbackslash{}n 3.\{$c^q_3$\} $\dots$} \\ \texttt{please write a correct answering passage.} \\
\midrule\midrule
Prompt for \textbf{scifact}. \\ \midrule
\texttt{Give a question ``\{$q$\}'' and its possible scientific paper passages (most of these passages are wrong) enumerated as: \textbackslash{}n~1.\{$c^q_1$\} \textbackslash{}n 2.\{$c^q_2$\} \textbackslash{}n 3.\{$c^q_3$\} $\dots$} \\ \texttt{please write a correct scientific paper passage.} \\
\midrule\midrule
Prompt for \textbf{arguana}. \\ \midrule
\texttt{Give a question ``\{$q$\}'' and its possible counter-argument passages (most of these passages are wrong) enumerated as: \textbackslash{}n~1.\{$c^q_1$\} \textbackslash{}n 2.\{$c^q_2$\} \textbackslash{}n 3.\{$c^q_3$\} $\dots$} \\ \texttt{please write a correct counter-argument passage.} \\
\midrule\midrule
Prompt for \textbf{trec-covid}. \\ \midrule
\texttt{Give a question ``\{$q$\}'' and its possible scientific paper passages (most of these passages are wrong) enumerated as: \textbackslash{}n~1.\{$c^q_1$\} \textbackslash{}n 2.\{$c^q_2$\} \textbackslash{}n 3.\{$c^q_3$\} $\dots$} \\ \texttt{please write a correct scientific paper passage.} \\
\midrule\midrule
Prompt for \textbf{fiqa}. \\ \midrule
\texttt{Give a question ``\{$q$\}'' and its possible answering financial article passages (most of these passages are wrong) enumerated as: \textbackslash{}n~1.\{$c^q_1$\} \textbackslash{}n 2.\{$c^q_2$\} \textbackslash{}n 3.\{$c^q_3$\} $\dots$} \\ \texttt{please write a correct answering financial article passage.} \\
\midrule\midrule
Prompt for \textbf{dbpedia}. \\ \midrule
\texttt{Give a question ``\{$q$\}'' and its possible answering passages (most of these passages are wrong) enumerated as: \textbackslash{}n~1.\{$c^q_1$\} \textbackslash{}n 2.\{$c^q_2$\} \textbackslash{}n 3.\{$c^q_3$\} $\dots$} \\ \texttt{please write a correct answering passage.} \\
\midrule\midrule
Prompt for \textbf{trec-news}. \\ \midrule
\texttt{Give a question ``\{$q$\}'' and its possible relevant passages (most of these passages are wrong) enumerated as: \textbackslash{}n~1.\{$c^q_1$\} \textbackslash{}n 2.\{$c^q_2$\} \textbackslash{}n 3.\{$c^q_3$\} $\dots$} \\ \texttt{please write a correct relevant passage.} \\
\bottomrule
\end{tabular}
\caption{Our prompts for all datasets. }
\label{tab:prompt_all}
\end{table*}

\end{document}

%% file: define_operator.tex
\usepackage{mathtools}
\usepackage{multirow}
\usepackage{multicol}
\usepackage{booktabs}
\usepackage{subfigure}

\DeclareMathOperator*{\enc}{Enc}

\DeclareMathOperator*{\relevancescore}{Rel}
\DeclareMathOperator*{\retriever}{Retriever}
\DeclareMathOperator*{\llm}{LLM}
\DeclareMathOperator*{\concat}{Concat}

%% file: math_commands.tex
\usepackage{amsmath,amsfonts,bm}

\def\eqref#1{equation~\ref{#1}}

\def\1{\bm{1}}

\def\vq{{\bm{q}}}

\DeclareMathAlphabet{\mathsfit}{\encodingdefault}{\sfdefault}{m}{sl}
\SetMathAlphabet{\mathsfit}{bold}{\encodingdefault}{\sfdefault}{bx}{n}

\def\sA{{\mathbb{A}}}

\def\sC{{\mathbb{C}}}
\def\sD{{\mathbb{D}}}










%% file: LameR-arxiv-v2.bbl
\begin{thebibliography}{45}
\providecommand{\natexlab}[1]{#1}
\providecommand{\url}[1]{\texttt{#1}}
\expandafter\ifx\csname urlstyle\endcsname\relax
  \providecommand{\doi}[1]{doi: #1}\else
  \providecommand{\doi}{doi: \begingroup \urlstyle{rm}\Url}\fi

\bibitem[Asai et~al.(2022)Asai, Schick, Lewis, Chen, Izacard, Riedel,
  Hajishirzi, and Yih]{Asai2022TaskRetrieval}
Akari Asai, Timo Schick, Patrick S.~H. Lewis, Xilun Chen, Gautier Izacard,
  Sebastian Riedel, Hannaneh Hajishirzi, and Wen{-}tau Yih.
\newblock Task-aware retrieval with instructions.
\newblock \emph{CoRR}, abs/2211.09260, 2022.
\newblock \doi{10.48550/arXiv.2211.09260}.
\newblock URL \url{https://doi.org/10.48550/arXiv.2211.09260}.

\bibitem[Boytsov et~al.(2023)Boytsov, Patel, Sourabh, Nisar, Kundu, Ramanathan,
  and Nyberg]{Boytsov2023InPars-Light}
Leonid Boytsov, Preksha Patel, Vivek Sourabh, Riddhi Nisar, Sayani Kundu, Ramya
  Ramanathan, and Eric Nyberg.
\newblock Inpars-light: Cost-effective unsupervised training of efficient
  rankers.
\newblock \emph{CoRR}, abs/2301.02998, 2023.
\newblock \doi{10.48550/arXiv.2301.02998}.
\newblock URL \url{https://doi.org/10.48550/arXiv.2301.02998}.

\bibitem[Brown et~al.(2020)Brown, Mann, Ryder, Subbiah, Kaplan, Dhariwal,
  Neelakantan, Shyam, Sastry, Askell, Agarwal, Herbert{-}Voss, Krueger,
  Henighan, Child, Ramesh, Ziegler, Wu, Winter, Hesse, Chen, Sigler, Litwin,
  Gray, Chess, Clark, Berner, McCandlish, Radford, Sutskever, and
  Amodei]{Brown2020GPT3}
Tom~B. Brown, Benjamin Mann, Nick Ryder, Melanie Subbiah, Jared Kaplan,
  Prafulla Dhariwal, Arvind Neelakantan, Pranav Shyam, Girish Sastry, Amanda
  Askell, Sandhini Agarwal, Ariel Herbert{-}Voss, Gretchen Krueger, Tom
  Henighan, Rewon Child, Aditya Ramesh, Daniel~M. Ziegler, Jeffrey Wu, Clemens
  Winter, Christopher Hesse, Mark Chen, Eric Sigler, Mateusz Litwin, Scott
  Gray, Benjamin Chess, Jack Clark, Christopher Berner, Sam McCandlish, Alec
  Radford, Ilya Sutskever, and Dario Amodei.
\newblock Language models are few-shot learners.
\newblock In Hugo Larochelle, Marc'Aurelio Ranzato, Raia Hadsell,
  Maria{-}Florina Balcan, and Hsuan{-}Tien Lin, editors, \emph{Advances in
  Neural Information Processing Systems 33: Annual Conference on Neural
  Information Processing Systems 2020, NeurIPS 2020, December 6-12, 2020,
  virtual}, 2020.
\newblock URL
  \url{https://proceedings.neurips.cc/paper/2020/hash/1457c0d6bfcb4967418bfb8ac142f64a-Abstract.html}.

\bibitem[Cai et~al.(2021)Cai, Fan, Guo, Sun, Zhang, and Cheng]{Cai2021IRSurvey}
Yinqiong Cai, Yixing Fan, Jiafeng Guo, Fei Sun, Ruqing Zhang, and Xueqi Cheng.
\newblock Semantic models for the first-stage retrieval: {A} comprehensive
  review.
\newblock \emph{CoRR}, abs/2103.04831, 2021.
\newblock URL \url{https://arxiv.org/abs/2103.04831}.

\bibitem[Chen et~al.(2017)Chen, Fisch, Weston, and Bordes]{Chen2017DrQA}
Danqi Chen, Adam Fisch, Jason Weston, and Antoine Bordes.
\newblock Reading wikipedia to answer open-domain questions.
\newblock In Regina Barzilay and Min{-}Yen Kan, editors, \emph{Proceedings of
  the 55th Annual Meeting of the Association for Computational Linguistics,
  {ACL} 2017, Vancouver, Canada, July 30 - August 4, Volume 1: Long Papers},
  pages 1870--1879. Association for Computational Linguistics, 2017.
\newblock \doi{10.18653/v1/P17-1171}.
\newblock URL \url{https://doi.org/10.18653/v1/P17-1171}.

\bibitem[Craswell et~al.(2020)Craswell, Mitra, Yilmaz, Campos, and
  Voorhees]{Craswell2020TREC19}
Nick Craswell, Bhaskar Mitra, Emine Yilmaz, Daniel Campos, and Ellen~M.
  Voorhees.
\newblock Overview of the {TREC} 2019 deep learning track.
\newblock \emph{CoRR}, abs/2003.07820, 2020.
\newblock URL \url{https://arxiv.org/abs/2003.07820}.

\bibitem[Craswell et~al.(2021)Craswell, Mitra, Yilmaz, and
  Campos]{Craswell2021TREC20}
Nick Craswell, Bhaskar Mitra, Emine Yilmaz, and Daniel Campos.
\newblock Overview of the {TREC} 2020 deep learning track.
\newblock \emph{CoRR}, abs/2102.07662, 2021.
\newblock URL \url{https://arxiv.org/abs/2102.07662}.

\bibitem[Dai et~al.(2022)Dai, Zhao, Ma, Luan, Ni, Lu, Bakalov, Guu, Hall, and
  Chang]{Dai2022Promptagator}
Zhuyun Dai, Vincent~Y. Zhao, Ji~Ma, Yi~Luan, Jianmo Ni, Jing Lu, Anton Bakalov,
  Kelvin Guu, Keith~B. Hall, and Ming{-}Wei Chang.
\newblock Promptagator: Few-shot dense retrieval from 8 examples.
\newblock \emph{CoRR}, abs/2209.11755, 2022.
\newblock \doi{10.48550/arXiv.2209.11755}.
\newblock URL \url{https://doi.org/10.48550/arXiv.2209.11755}.

\bibitem[Devlin et~al.(2019)Devlin, Chang, Lee, and Toutanova]{Devlin2019BERT}
Jacob Devlin, Ming{-}Wei Chang, Kenton Lee, and Kristina Toutanova.
\newblock {BERT:} pre-training of deep bidirectional transformers for language
  understanding.
\newblock In Jill Burstein, Christy Doran, and Thamar Solorio, editors,
  \emph{Proceedings of the 2019 Conference of the North American Chapter of the
  Association for Computational Linguistics: Human Language Technologies,
  {NAACL-HLT} 2019, Minneapolis, MN, USA, June 2-7, 2019, Volume 1 (Long and
  Short Papers)}, pages 4171--4186. Association for Computational Linguistics,
  2019.
\newblock \doi{10.18653/v1/n19-1423}.
\newblock URL \url{https://doi.org/10.18653/v1/n19-1423}.

\bibitem[Dua et~al.(2022)Dua, Strubell, Singh, and Verga]{Dua2022AdaptAnnotate}
Dheeru Dua, Emma Strubell, Sameer Singh, and Pat Verga.
\newblock To adapt or to annotate: Challenges and interventions for domain
  adaptation in open-domain question answering.
\newblock \emph{CoRR}, abs/2212.10381, 2022.
\newblock \doi{10.48550/arXiv.2212.10381}.
\newblock URL \url{https://doi.org/10.48550/arXiv.2212.10381}.

\bibitem[Gao and Callan(2022)]{Gao2022Reranking}
Luyu Gao and Jamie Callan.
\newblock Long document re-ranking with modular re-ranker.
\newblock In Enrique Amig{\'{o}}, Pablo Castells, Julio Gonzalo, Ben
  Carterette, J.~Shane Culpepper, and Gabriella Kazai, editors, \emph{{SIGIR}
  '22: The 45th International {ACM} {SIGIR} Conference on Research and
  Development in Information Retrieval, Madrid, Spain, July 11 - 15, 2022},
  pages 2371--2376. {ACM}, 2022.
\newblock \doi{10.1145/3477495.3531860}.
\newblock URL \url{https://doi.org/10.1145/3477495.3531860}.

\bibitem[Gao et~al.(2022)Gao, Ma, Lin, and Callan]{Gao2022hyde}
Luyu Gao, Xueguang Ma, Jimmy Lin, and Jamie Callan.
\newblock Precise zero-shot dense retrieval without relevance labels.
\newblock \emph{CoRR}, abs/2212.10496, 2022.
\newblock \doi{10.48550/arXiv.2212.10496}.
\newblock URL \url{https://doi.org/10.48550/arXiv.2212.10496}.

\bibitem[Guu et~al.(2020)Guu, Lee, Tung, Pasupat, and Chang]{Guu2020RetriAugLM}
Kelvin Guu, Kenton Lee, Zora Tung, Panupong Pasupat, and Ming{-}Wei Chang.
\newblock {REALM:} retrieval-augmented language model pre-training.
\newblock \emph{CoRR}, abs/2002.08909, 2020.
\newblock URL \url{https://arxiv.org/abs/2002.08909}.

\bibitem[He et~al.(2023)He, Zhang, and Roth]{He2023Rethinking}
Hangfeng He, Hongming Zhang, and Dan Roth.
\newblock Rethinking with retrieval: Faithful large language model inference.
\newblock \emph{CoRR}, abs/2301.00303, 2023.
\newblock \doi{10.48550/arXiv.2301.00303}.
\newblock URL \url{https://doi.org/10.48550/arXiv.2301.00303}.

\bibitem[Izacard et~al.(2021)Izacard, Caron, Hosseini, Riedel, Bojanowski,
  Joulin, and Grave]{Izacard2021Contriever}
Gautier Izacard, Mathilde Caron, Lucas Hosseini, Sebastian Riedel, Piotr
  Bojanowski, Armand Joulin, and Edouard Grave.
\newblock Towards unsupervised dense information retrieval with contrastive
  learning.
\newblock \emph{CoRR}, abs/2112.09118, 2021.
\newblock URL \url{https://arxiv.org/abs/2112.09118}.

\bibitem[Jeronymo et~al.(2023)Jeronymo, Bonifacio, Abonizio, Fadaee,
  de~Alencar~Lotufo, Zavrel, and Nogueira]{Jeronymo2023InPars-v2}
Vitor Jeronymo, Luiz~Henrique Bonifacio, Hugo Abonizio, Marzieh Fadaee, Roberto
  de~Alencar~Lotufo, Jakub Zavrel, and Rodrigo~Frassetto Nogueira.
\newblock Inpars-v2: Large language models as efficient dataset generators for
  information retrieval.
\newblock \emph{CoRR}, abs/2301.01820, 2023.
\newblock \doi{10.48550/arXiv.2301.01820}.
\newblock URL \url{https://doi.org/10.48550/arXiv.2301.01820}.

\bibitem[Karpukhin et~al.(2020)Karpukhin, Oguz, Min, Lewis, Wu, Edunov, Chen,
  and Yih]{Karpukhin2020DPR}
Vladimir Karpukhin, Barlas Oguz, Sewon Min, Patrick S.~H. Lewis, Ledell Wu,
  Sergey Edunov, Danqi Chen, and Wen{-}tau Yih.
\newblock Dense passage retrieval for open-domain question answering.
\newblock In Bonnie Webber, Trevor Cohn, Yulan He, and Yang Liu, editors,
  \emph{Proceedings of the 2020 Conference on Empirical Methods in Natural
  Language Processing, {EMNLP} 2020, Online, November 16-20, 2020}, pages
  6769--6781. Association for Computational Linguistics, 2020.
\newblock \doi{10.18653/v1/2020.emnlp-main.550}.
\newblock URL \url{https://doi.org/10.18653/v1/2020.emnlp-main.550}.

\bibitem[Lee et~al.(2019)Lee, Chang, and Toutanova]{Lee2019ICT}
Kenton Lee, Ming{-}Wei Chang, and Kristina Toutanova.
\newblock Latent retrieval for weakly supervised open domain question
  answering.
\newblock In Anna Korhonen, David~R. Traum, and Llu{\'{\i}}s M{\`{a}}rquez,
  editors, \emph{Proceedings of the 57th Conference of the Association for
  Computational Linguistics, {ACL} 2019, Florence, Italy, July 28- August 2,
  2019, Volume 1: Long Papers}, pages 6086--6096. Association for Computational
  Linguistics, 2019.
\newblock \doi{10.18653/v1/p19-1612}.
\newblock URL \url{https://doi.org/10.18653/v1/p19-1612}.

\bibitem[Lin et~al.(2021)Lin, Ma, Lin, Yang, Pradeep, and
  Nogueira]{Lin2021pyserini}
Jimmy Lin, Xueguang Ma, Sheng-Chieh Lin, Jheng-Hong Yang, Ronak Pradeep, and
  Rodrigo Nogueira.
\newblock {Pyserini}: A {Python} toolkit for reproducible information retrieval
  research with sparse and dense representations.
\newblock In \emph{Proceedings of the 44th Annual International ACM SIGIR
  Conference on Research and Development in Information Retrieval (SIGIR
  2021)}, pages 2356--2362, 2021.

\bibitem[Liu et~al.(2022)Liu, Shen, Zhang, Dolan, Carin, and
  Chen]{Liu2022ICLWhat}
Jiachang Liu, Dinghan Shen, Yizhe Zhang, Bill Dolan, Lawrence Carin, and Weizhu
  Chen.
\newblock What makes good in-context examples for gpt-3?
\newblock In Eneko Agirre, Marianna Apidianaki, and Ivan Vulic, editors,
  \emph{Proceedings of Deep Learning Inside Out: The 3rd Workshop on Knowledge
  Extraction and Integration for Deep Learning Architectures, DeeLIO@ACL 2022,
  Dublin, Ireland and Online, May 27, 2022}, pages 100--114. Association for
  Computational Linguistics, 2022.
\newblock \doi{10.18653/v1/2022.deelio-1.10}.
\newblock URL \url{https://doi.org/10.18653/v1/2022.deelio-1.10}.

\bibitem[Lyu et~al.(2022)Lyu, Min, Beltagy, Zettlemoyer, and
  Hajishirzi]{Lyu2022ZICL}
Xinxi Lyu, Sewon Min, Iz~Beltagy, Luke Zettlemoyer, and Hannaneh Hajishirzi.
\newblock {Z-ICL:} zero-shot in-context learning with pseudo-demonstrations.
\newblock \emph{CoRR}, abs/2212.09865, 2022.
\newblock \doi{10.48550/arXiv.2212.09865}.
\newblock URL \url{https://doi.org/10.48550/arXiv.2212.09865}.

\bibitem[Mikolov et~al.(2013)Mikolov, Chen, Corrado, and
  Dean]{Mikolov2013WordEmb}
Tom{\'{a}}s Mikolov, Kai Chen, Greg Corrado, and Jeffrey Dean.
\newblock Efficient estimation of word representations in vector space.
\newblock In Yoshua Bengio and Yann LeCun, editors, \emph{1st International
  Conference on Learning Representations, {ICLR} 2013, Scottsdale, Arizona,
  USA, May 2-4, 2013, Workshop Track Proceedings}, 2013.
\newblock URL \url{http://arxiv.org/abs/1301.3781}.

\bibitem[Min et~al.(2022)Min, Lyu, Holtzman, Artetxe, Lewis, Hajishirzi, and
  Zettlemoyer]{Min2022ICLRethinking}
Sewon Min, Xinxi Lyu, Ari Holtzman, Mikel Artetxe, Mike Lewis, Hannaneh
  Hajishirzi, and Luke Zettlemoyer.
\newblock Rethinking the role of demonstrations: What makes in-context learning
  work?
\newblock In Yoav Goldberg, Zornitsa Kozareva, and Yue Zhang, editors,
  \emph{Proceedings of the 2022 Conference on Empirical Methods in Natural
  Language Processing, {EMNLP} 2022, Abu Dhabi, United Arab Emirates, December
  7-11, 2022}, pages 11048--11064. Association for Computational Linguistics,
  2022.
\newblock URL \url{https://aclanthology.org/2022.emnlp-main.759}.

\bibitem[Muennighoff(2022)]{Muennighoff2022SGPT}
Niklas Muennighoff.
\newblock {SGPT:} {GPT} sentence embeddings for semantic search.
\newblock \emph{CoRR}, abs/2202.08904, 2022.
\newblock URL \url{https://arxiv.org/abs/2202.08904}.

\bibitem[Nguyen et~al.(2016)Nguyen, Rosenberg, Song, Gao, Tiwary, Majumder, and
  Deng]{Nguyen2016MSMARCO}
Tri Nguyen, Mir Rosenberg, Xia Song, Jianfeng Gao, Saurabh Tiwary, Rangan
  Majumder, and Li~Deng.
\newblock {MS} {MARCO:} {A} human generated machine reading comprehension
  dataset.
\newblock In Tarek~Richard Besold, Antoine Bordes, Artur~S. d'Avila Garcez, and
  Greg Wayne, editors, \emph{Proceedings of the Workshop on Cognitive
  Computation: Integrating neural and symbolic approaches 2016 co-located with
  the 30th Annual Conference on Neural Information Processing Systems {(NIPS}
  2016), Barcelona, Spain, December 9, 2016}, volume 1773 of \emph{{CEUR}
  Workshop Proceedings}. CEUR-WS.org, 2016.
\newblock URL \url{http://ceur-ws.org/Vol-1773/CoCoNIPS\_2016\_paper9.pdf}.

\bibitem[Ni et~al.(2021)Ni, Qu, Lu, Dai, {\'{A}}brego, Ma, Zhao, Luan, Hall,
  Chang, and Yang]{Ni2021GTR}
Jianmo Ni, Chen Qu, Jing Lu, Zhuyun Dai, Gustavo~Hern{\'{a}}ndez {\'{A}}brego,
  Ji~Ma, Vincent~Y. Zhao, Yi~Luan, Keith~B. Hall, Ming{-}Wei Chang, and Yinfei
  Yang.
\newblock Large dual encoders are generalizable retrievers.
\newblock \emph{CoRR}, abs/2112.07899, 2021.
\newblock URL \url{https://arxiv.org/abs/2112.07899}.

\bibitem[Razeghi et~al.(2022)Razeghi, IV, Gardner, and
  Singh]{Razeghi2022ICLFreq}
Yasaman Razeghi, Robert L.~Logan IV, Matt Gardner, and Sameer Singh.
\newblock Impact of pretraining term frequencies on few-shot numerical
  reasoning.
\newblock In Yoav Goldberg, Zornitsa Kozareva, and Yue Zhang, editors,
  \emph{Findings of the Association for Computational Linguistics: {EMNLP}
  2022, Abu Dhabi, United Arab Emirates, December 7-11, 2022}, pages 840--854.
  Association for Computational Linguistics, 2022.
\newblock URL \url{https://aclanthology.org/2022.findings-emnlp.59}.

\bibitem[Robertson and Zaragoza(2009)]{Robertson2009BM25}
Stephen~E. Robertson and Hugo Zaragoza.
\newblock The probabilistic relevance framework: {BM25} and beyond.
\newblock \emph{Found. Trends Inf. Retr.}, 3\penalty0 (4):\penalty0 333--389,
  2009.
\newblock \doi{10.1561/1500000019}.
\newblock URL \url{https://doi.org/10.1561/1500000019}.

\bibitem[Rubin et~al.(2022)Rubin, Herzig, and Berant]{Rubin2022ICLL2retrieval}
Ohad Rubin, Jonathan Herzig, and Jonathan Berant.
\newblock Learning to retrieve prompts for in-context learning.
\newblock In Marine Carpuat, Marie{-}Catherine de~Marneffe, and Iv{\'{a}}n
  Vladimir~Meza Ru{\'{\i}}z, editors, \emph{Proceedings of the 2022 Conference
  of the North American Chapter of the Association for Computational
  Linguistics: Human Language Technologies, {NAACL} 2022, Seattle, WA, United
  States, July 10-15, 2022}, pages 2655--2671. Association for Computational
  Linguistics, 2022.
\newblock \doi{10.18653/v1/2022.naacl-main.191}.
\newblock URL \url{https://doi.org/10.18653/v1/2022.naacl-main.191}.

\bibitem[Saad{-}Falcon et~al.(2023)Saad{-}Falcon, Khattab, Santhanam, Florian,
  Franz, Roukos, Sil, Sultan, and Potts]{Falcon2023UDAPDR}
Jon Saad{-}Falcon, Omar Khattab, Keshav Santhanam, Radu Florian, Martin Franz,
  Salim Roukos, Avirup Sil, Md.~Arafat Sultan, and Christopher Potts.
\newblock {UDAPDR:} unsupervised domain adaptation via {LLM} prompting and
  distillation of rerankers.
\newblock \emph{CoRR}, abs/2303.00807, 2023.
\newblock \doi{10.48550/arXiv.2303.00807}.
\newblock URL \url{https://doi.org/10.48550/arXiv.2303.00807}.

\bibitem[Shen et~al.(2022)Shen, Geng, Tao, Xu, Huang, Jiao, Yang, and
  Jiang]{Shen2023LexMAE}
Tao Shen, Xiubo Geng, Chongyang Tao, Can Xu, Xiaolong Huang, Binxing Jiao,
  Linjun Yang, and Daxin Jiang.
\newblock Lexmae: Lexicon-bottlenecked pretraining for large-scale retrieval.
\newblock \emph{CoRR}, abs/2208.14754, 2022.
\newblock \doi{10.48550/arXiv.2208.14754}.
\newblock URL \url{https://doi.org/10.48550/arXiv.2208.14754}.

\bibitem[Shuster et~al.(2021)Shuster, Poff, Chen, Kiela, and
  Weston]{Shuster2021RetrievalHalluci}
Kurt Shuster, Spencer Poff, Moya Chen, Douwe Kiela, and Jason Weston.
\newblock Retrieval augmentation reduces hallucination in conversation.
\newblock In Marie{-}Francine Moens, Xuanjing Huang, Lucia Specia, and
  Scott~Wen{-}tau Yih, editors, \emph{Findings of the Association for
  Computational Linguistics: {EMNLP} 2021, Virtual Event / Punta Cana,
  Dominican Republic, 16-20 November, 2021}, pages 3784--3803. Association for
  Computational Linguistics, 2021.
\newblock \doi{10.18653/v1/2021.findings-emnlp.320}.
\newblock URL \url{https://doi.org/10.18653/v1/2021.findings-emnlp.320}.

\bibitem[Taori et~al.(2023)Taori, Gulrajani, Zhang, Dubois, Li, Guestrin,
  Liang, and Hashimoto]{Taori2023alpaca}
Rohan Taori, Ishaan Gulrajani, Tianyi Zhang, Yann Dubois, Xuechen Li, Carlos
  Guestrin, Percy Liang, and Tatsunori~B. Hashimoto.
\newblock Stanford alpaca: An instruction-following llama model.
\newblock \url{https://github.com/tatsu-lab/stanford_alpaca}, 2023.

\bibitem[Thakur et~al.(2021)Thakur, Reimers, R{\"{u}}ckl{\'{e}}, Srivastava,
  and Gurevych]{Thakur2021BEIR}
Nandan Thakur, Nils Reimers, Andreas R{\"{u}}ckl{\'{e}}, Abhishek Srivastava,
  and Iryna Gurevych.
\newblock {BEIR:} {A} heterogenous benchmark for zero-shot evaluation of
  information retrieval models.
\newblock \emph{CoRR}, abs/2104.08663, 2021.
\newblock URL \url{https://arxiv.org/abs/2104.08663}.

\bibitem[Touvron et~al.(2023)Touvron, Lavril, Izacard, Martinet, Lachaux,
  Lacroix, Rozi{\`{e}}re, Goyal, Hambro, Azhar, Rodriguez, Joulin, Grave, and
  Lample]{Touvron2023llama}
Hugo Touvron, Thibaut Lavril, Gautier Izacard, Xavier Martinet, Marie{-}Anne
  Lachaux, Timoth{\'{e}}e Lacroix, Baptiste Rozi{\`{e}}re, Naman Goyal, Eric
  Hambro, Faisal Azhar, Aur{\'{e}}lien Rodriguez, Armand Joulin, Edouard Grave,
  and Guillaume Lample.
\newblock Llama: Open and efficient foundation language models.
\newblock \emph{CoRR}, abs/2302.13971, 2023.
\newblock \doi{10.48550/arXiv.2302.13971}.
\newblock URL \url{https://doi.org/10.48550/arXiv.2302.13971}.

\bibitem[Trivedi et~al.(2022)Trivedi, Balasubramanian, Khot, and
  Sabharwal]{Trivedi2022CotRetri}
Harsh Trivedi, Niranjan Balasubramanian, Tushar Khot, and Ashish Sabharwal.
\newblock Interleaving retrieval with chain-of-thought reasoning for
  knowledge-intensive multi-step questions.
\newblock \emph{CoRR}, abs/2212.10509, 2022.
\newblock \doi{10.48550/arXiv.2212.10509}.
\newblock URL \url{https://doi.org/10.48550/arXiv.2212.10509}.

\bibitem[Wang et~al.(2022{\natexlab{a}})Wang, Yang, Huang, Jiao, Yang, Jiang,
  Majumder, and Wei]{Wang2022e5}
Liang Wang, Nan Yang, Xiaolong Huang, Binxing Jiao, Linjun Yang, Daxin Jiang,
  Rangan Majumder, and Furu Wei.
\newblock Text embeddings by weakly-supervised contrastive pre-training.
\newblock \emph{CoRR}, abs/2212.03533, 2022{\natexlab{a}}.
\newblock \doi{10.48550/arXiv.2212.03533}.
\newblock URL \url{https://doi.org/10.48550/arXiv.2212.03533}.

\bibitem[Wang et~al.(2022{\natexlab{b}})Wang, Yang, Huang, Jiao, Yang, Jiang,
  Majumder, and Wei]{Wang2022simlm}
Liang Wang, Nan Yang, Xiaolong Huang, Binxing Jiao, Linjun Yang, Daxin Jiang,
  Rangan Majumder, and Furu Wei.
\newblock Simlm: Pre-training with representation bottleneck for dense passage
  retrieval.
\newblock \emph{CoRR}, abs/2207.02578, 2022{\natexlab{b}}.
\newblock \doi{10.48550/arXiv.2207.02578}.
\newblock URL \url{https://doi.org/10.48550/arXiv.2207.02578}.

\bibitem[Wang et~al.(2023)Wang, Yang, and Wei]{Wang2023Query2doc}
Liang Wang, Nan Yang, and Furu Wei.
\newblock Query2doc: Query expansion with large language models.
\newblock \emph{CoRR}, abs/2303.07678, 2023.
\newblock \doi{10.48550/arXiv.2303.07678}.
\newblock URL \url{https://doi.org/10.48550/arXiv.2303.07678}.

\bibitem[Xie et~al.(2022)Xie, Raghunathan, Liang, and Ma]{Xie2022ICLBayesian}
Sang~Michael Xie, Aditi Raghunathan, Percy Liang, and Tengyu Ma.
\newblock An explanation of in-context learning as implicit bayesian inference.
\newblock In \emph{The Tenth International Conference on Learning
  Representations, {ICLR} 2022, Virtual Event, April 25-29, 2022}.
  OpenReview.net, 2022.
\newblock URL \url{https://openreview.net/forum?id=RdJVFCHjUMI}.

\bibitem[Xiong et~al.(2021)Xiong, Xiong, Li, Tang, Liu, Bennett, Ahmed, and
  Overwijk]{Xiong2021ANCE}
Lee Xiong, Chenyan Xiong, Ye~Li, Kwok{-}Fung Tang, Jialin Liu, Paul~N. Bennett,
  Junaid Ahmed, and Arnold Overwijk.
\newblock Approximate nearest neighbor negative contrastive learning for dense
  text retrieval.
\newblock In \emph{9th International Conference on Learning Representations,
  {ICLR} 2021, Virtual Event, Austria, May 3-7, 2021}. OpenReview.net, 2021.
\newblock URL \url{https://openreview.net/forum?id=zeFrfgyZln}.

\bibitem[Yu et~al.(2022)Yu, Iter, Wang, Xu, Ju, Sanyal, Zhu, Zeng, and
  Jiang]{Yu2022GenerateRetrieve}
Wenhao Yu, Dan Iter, Shuohang Wang, Yichong Xu, Mingxuan Ju, Soumya Sanyal,
  Chenguang Zhu, Michael Zeng, and Meng Jiang.
\newblock Generate rather than retrieve: Large language models are strong
  context generators.
\newblock \emph{CoRR}, abs/2209.10063, 2022.
\newblock \doi{10.48550/arXiv.2209.10063}.
\newblock URL \url{https://doi.org/10.48550/arXiv.2209.10063}.

\bibitem[Zhao et~al.(2020)Zhao, Wu, Xu, Tao, Zhao, and Yan]{Zhao2020KGC}
Xueliang Zhao, Wei Wu, Can Xu, Chongyang Tao, Dongyan Zhao, and Rui Yan.
\newblock Knowledge-grounded dialogue generation with pre-trained language
  models.
\newblock In Bonnie Webber, Trevor Cohn, Yulan He, and Yang Liu, editors,
  \emph{Proceedings of the 2020 Conference on Empirical Methods in Natural
  Language Processing, {EMNLP} 2020, Online, November 16-20, 2020}, pages
  3377--3390. Association for Computational Linguistics, 2020.
\newblock \doi{10.18653/v1/2020.emnlp-main.272}.
\newblock URL \url{https://doi.org/10.18653/v1/2020.emnlp-main.272}.

\bibitem[Zhou et~al.(2022{\natexlab{a}})Zhou, Li, Shang, Luo, Zhan, Hu, Zhang,
  Jiang, Cao, Yu, Jiang, Liu, and Chen]{Zhou2022hyperlink}
Jiawei Zhou, Xiaoguang Li, Lifeng Shang, Lan Luo, Ke~Zhan, Enrui Hu, Xinyu
  Zhang, Hao Jiang, Zhao Cao, Fan Yu, Xin Jiang, Qun Liu, and Lei Chen.
\newblock Hyperlink-induced pre-training for passage retrieval in open-domain
  question answering.
\newblock In Smaranda Muresan, Preslav Nakov, and Aline Villavicencio, editors,
  \emph{Proceedings of the 60th Annual Meeting of the Association for
  Computational Linguistics (Volume 1: Long Papers), {ACL} 2022, Dublin,
  Ireland, May 22-27, 2022}, pages 7135--7146. Association for Computational
  Linguistics, 2022{\natexlab{a}}.
\newblock \doi{10.18653/v1/2022.acl-long.493}.
\newblock URL \url{https://doi.org/10.18653/v1/2022.acl-long.493}.

\bibitem[Zhou et~al.(2022{\natexlab{b}})Zhou, Shen, Geng, Tao, Xu, Long, Jiao,
  and Jiang]{Zhou2022r2anker}
Yucheng Zhou, Tao Shen, Xiubo Geng, Chongyang Tao, Can Xu, Guodong Long,
  Binxing Jiao, and Daxin Jiang.
\newblock Towards robust ranker for text retrieval.
\newblock \emph{CoRR}, abs/2206.08063, 2022{\natexlab{b}}.
\newblock \doi{10.48550/arXiv.2206.08063}.
\newblock URL \url{https://doi.org/10.48550/arXiv.2206.08063}.

\end{thebibliography}
